\documentclass[letterpaper,10pt,conference]{IEEEtran}
\IEEEoverridecommandlockouts

\usepackage{cite}
\usepackage[T1]{fontenc}
\usepackage{lmodern} 
\usepackage{amsmath,amssymb,amsfonts}
\usepackage{graphicx}
\usepackage{booktabs}
\usepackage{algorithm}
\usepackage{algorithmic}
\usepackage{textcomp}
\usepackage{xcolor}
\usepackage{balance}
\usepackage{multirow}
\usepackage{caption}
\usepackage{subcaption}
\usepackage{verbatim}

\def\BibTeX{{\rm B\kern-.05em{\sc i\kern-.025em b}\kern-.08em
    T\kern-.1667em\lower.7ex\hbox{E}\kern-.125emX}}
\begin{document}

\title{Cross-household Transfer Learning Approach with LSTM-based Demand Forecasting}                     

\author{\IEEEauthorblockN{Manal Rahal}
\IEEEauthorblockA{\textit{Department of Mathematics and Computer Science} \\
\textit{Karlstad University, Sweden}}
\IEEEauthorblockA{\textit{Department of Research and Development} \\
\textit{Thermia AB, Sweden}\\
manal.rahal@kau.se}
\and
\IEEEauthorblockN{Bestoun S. Ahmed}
\IEEEauthorblockA{\textit{Department of Mathematics and Computer Science} \\
\textit{Karlstad University, Sweden}\\
\textit{American University of Bahrain, Bahrain}\\
bestoun@kau.se}
\and
\IEEEauthorblockN{Roger Renström}
\IEEEauthorblockA{\textit{Department of Mathematics and Computer Science} \\
\textit{Karlstad University, Sweden}\\
roger.renstrom@kau.se}
\and
\IEEEauthorblockN{Robert Stener}
\IEEEauthorblockA{\textit{Department of Research and Development} \\
\textit{Thermia AB, Sweden}\\
robert.stener@thermia.com}
}

\maketitle

\begin{abstract}
With the rapid increase in residential heat pump (HP) installations, optimizing hot water production in households is essential, yet it is faced with major technical and scalability challenges. Adapting production to actual household needs necessitates accurate forecasting of hot water demand to ensure comfort and, most importantly, to reduce energy waste. However, the conventional approach of training separate machine learning models for each household becomes computationally expensive at scale, particularly in cloud-connected HP deployments.

This study introduces DELTAiF, a transfer learning (TL)-based framework that offers scalable and accurate prediction of households' consumption of hot water. By predicting large usage of hot water, such as in showers, DELTAiF enables adaptive yet scalable hot water production at the household level. DELTAiF leverages learned knowledge from a representative household and fine-tunes it across others, eliminating the need to train separate machine learning models for each HP installation. This approach reduces overall training time by approximately 67\% while maintaining high predictive accuracy values between 0.874–0.991 and mean absolute percentage error values of 0.001-0.017. The results show that TL is particularly effective when the source household exhibits regular consumption patterns, hence enabling hot water demand forecasting at scale.

\end{abstract}


\section{Introduction}
Energy forecasting has become increasingly essential in the modern energy landscape as consumers seek to reduce costs and minimize environmental impact. In residential settings, where heat pumps (HPs) are being adopted at unprecedented rates, accurate energy demand predictions play an important role in optimizing resource allocation, enhancing infrastructure planning, and facilitating the development of intelligent energy systems \cite{Ukoba2024}. 

Despite the advanced technologies to collect big data from a real HP installation, using machine learning (ML) to improve energy-related tasks remains a significant challenge due to the complexity and scale of the data involved \cite{FAN2020}. This challenge is particularly acute in residential HP deployments, where each installation generates thousands of data points daily, requiring efficient processing and analysis methods. Furthermore, the diversity in household consumption patterns limits the direct adaptation of ML models across multiple household datasets. As a result, ML models can underperform, suffer from long training times, or lack sufficient data to generalize well \cite{Himeur2022}.

To address these gaps, we propose a DELTAiF, a transfer learning (TL) based approach that efficiently scales across multiple households while ensuring high prediction accuracy. This is achieved by training a Long Short-Term Memory (LSTM) base model on a representative household and fine-tuning it for the other individual installations. Our approach builds on the strengths of both TL and LSTM while reducing the computational burden of training individual models for each HP installation. To this end, the primary contributions of the paper include:

\begin{enumerate}
    \item  We propose DELTAiF (Demand Estimation using LSTM, Transfer Learning, and Isolation Forest) framework that integrates LSTM with TL and the isolation forest (iForest) to forecast and steer hot water production in an HP. The framework reduces computational overhead by 67\% compared to training individual models while maintaining high prediction accuracy.
    
    \item  We implement cross-household TL analysis to identify the optimal source domain for TL in HP systems. 
    
    \item To demonstrate the effectiveness of DELTAiF, we implement it on six Swedish household datasets collected from real HP installations. Three conventional forecasting evaluation metrics, coefficient of determination (\( R^2 \)), root mean squared error (RMSE), and mean absolute percentage error (MAPE), are analyzed and compared. 
    
    \item We introduce an automated pipeline to generate a weekly hot water demand calendar, enabling adaptive hot water production that meets the actual consumption needs of a household.
\end{enumerate}


\section{Background and Related Work}\label{sec:background}
This section introduces TL and demonstrates its value in relevant applications and domains. Additionally, provides context for residential HP operations and the challenges encountered in optimizing hot water production.

\subsection{Transfer Learning}
TL has emerged as a promising practice in ML, particularly in addressing data scarcity limitations and reducing high training costs, mainly in the fields of computer vision and natural language processing (NLP) \cite{Neyshabur2020}. The core concept of TL is to enhance learning in a target domain by utilizing knowledge learned from a related source domain \cite{Zhuang2021}. 

In practice, the transfer approach uses a pretrained model, where the top layers related to specific tasks are removed, and the bottom layers are fine-tuned for the target task \cite{You2020}. Neyshabur \emph{et al.} \cite{Neyshabur2020} observed that the lower layers of a model handle general features, while the higher layers are more sensitive to changes in their parameters. Typically, in the absence of TL, a deep learning model initializes with random weights, which are gradually adjusted during the training phase. Compared to beginning with randomly initialized weights, starting with weights from a pretrained network, even with an unrelated dataset, substantially boosts training efficiency for new tasks \cite{Yosinski2014}.

While TL has been extensively applied in fields such as medicine \cite{Maqsood2019, Shin2016}, transportation \cite{CHEN2023}, and recommendation systems \cite{Yuan2019, Wang2024book}, it has also found its way into the energy sector, where it is increasingly combined with deep learning models like LSTM to improve forecasting performance. For example, Kim \emph{et al.} \cite{Kim2023} integrates LSTM with TL to forecast building energy loads, under varying weather conditions. Similarly, \cite{YUAN2023} demonstrated that the attention mechanism-based TL model outperforms traditional models such as auto-regression in 24-hour day-ahead prediction tasks. In another study, Zhou \emph{et al.} \cite{ZHOU2023} integrate LSTM and TL to accurately predict the electric power consumption of a target building where historical data are scarce. In a related context, Liu \emph{et al.} \cite{LIU2022} leverages TL and fine-tuning to train error models with small training datasets.

In the context of household-level energy systems, TL offers distinct advantages by using patterns from well-understood households to improve predictions for new installations or households with limited historical data. However, ML applications in this area are limited to performance monitoring and fault detection in large installations \cite{Song2023consumption}, such as in \cite{zhu2021, Sunal2024}. Therefore, accurate demand forecasting combined with TL can offer an opportunity for demand-driven solutions that translate into substantial cost savings \cite{Mathumitha2024}.  

\subsection{Hot water production in a residential HP}
HP stands out among other heating alternatives as a clean and renewable energy source, providing heating in the winter and cooling in the summer for residential buildings \cite{Adebayo2024}. In addition to heating, an HP manages domestic hot water production through a temperature-controlled tank: heated water naturally rises to the top for immediate use, while cold water fills the tank from the bottom.

A key challenge lies in ensuring the continuous availability of hot water that meets household demand throughout the year, while optimizing production efficiency to minimize energy losses to the surroundings during idle times. This task becomes particularly complex in residential contexts, where consumption patterns vary widely both within and across households. As a result, household-specific control strategies are needed to sustain user comfort while maximizing energy efficiency.

Adaptive production of hot water is enabled by the smart sensors installed in the tank. The mid-tank sensor records temperature measurements at the middle of the tank ($t_\mathrm{mid}$), while the top-tank sensor records temperature measurements at the top ($t_\mathrm{top}$). During high consumption periods, $t_\mathrm{mid}$ drops significantly while $t_\mathrm{top}$ remains relatively stable, ensuring consistent hot water availability for users. The existing control system initiates production based on preset thresholds using the average of $t_\mathrm{mid}$ and $t_\mathrm{top}$. This threshold-based approach has a critical limitation; it may trigger production during low-demand periods, such as after midnight, due to natural temperature decline. This leads to unnecessary energy losses to the surroundings and inefficient production of hot water that is poorly synchronized with actual household demand.

Traditional ML approaches often fail to capture individual household characteristics due to the diverse nature of consumption patterns and socioeconomic factors \cite{tang2022machine}. However, combining LSTM and TL presents an opportunity to address this limitation by leveraging knowledge from pretrained LSTM models in similar time-series prediction tasks.

\section{DELTAiF Framework}\label{sec:framework}

In this paper, we propose DELTAiF, a TL-based framework, to predict consumption demand trends for hot water in households. The framework addresses two key challenges in residential HP deployments: prediction accuracy and computational scalability. Given the size of the train data, DELTAiF reduces per-household training time from 9 minutes to approximately 3 minutes while maintaining \( R^2 \) values between 0.874-0.991 across diverse household patterns. 

\begin{figure*}
    \centering
    \includegraphics[width=0.7\linewidth]{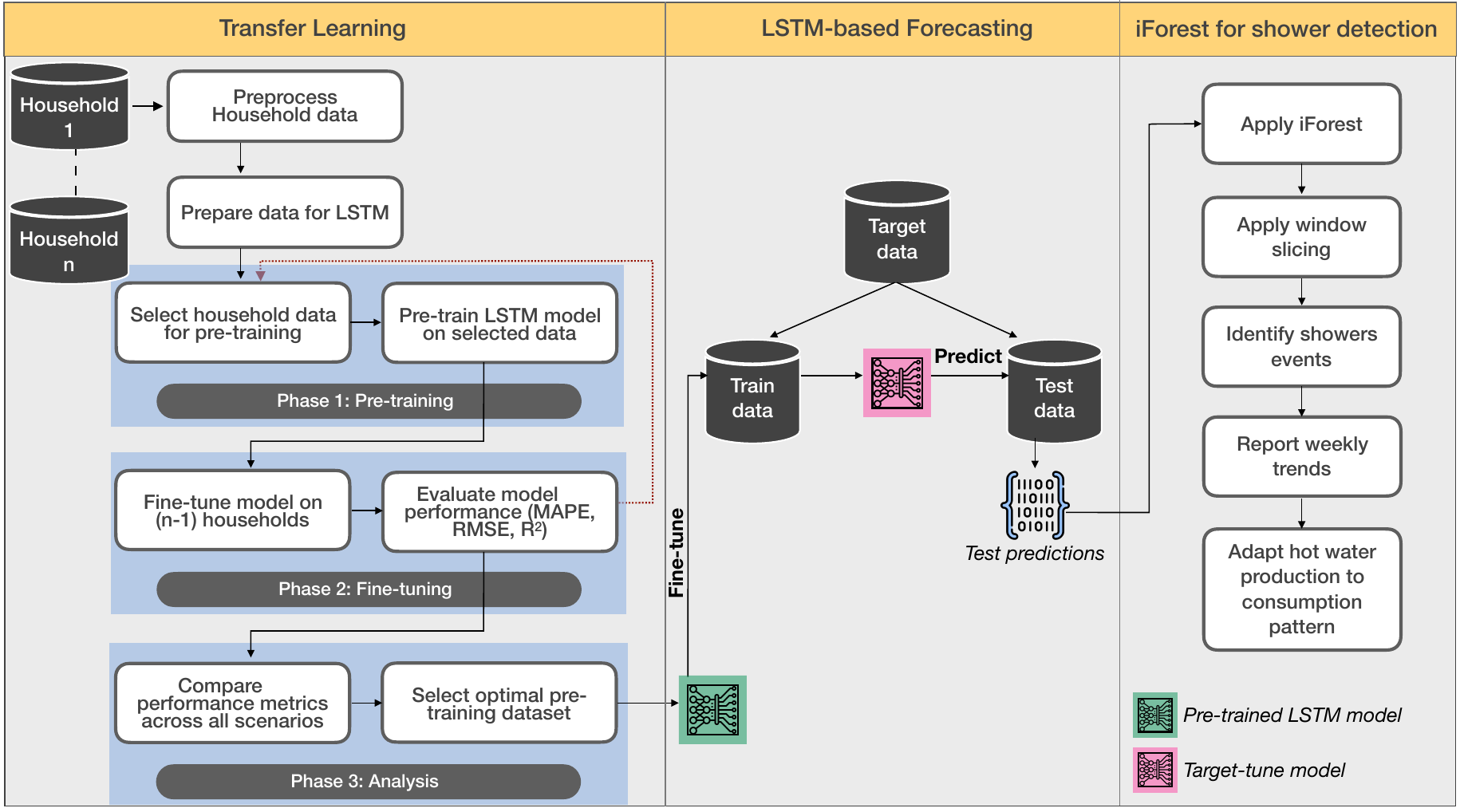}
    \caption{DELTAiF framework}
    \label{fig:Delta}
\end{figure*}

The framework shown in Figure \ref{fig:Delta} leverages the advantages of TL and integrates it with LSTM and iForest models in a sequential three-phase pipeline: (1) source domain selection and pre-training, (2) transfer and adaptation to target domains, and (3) event pattern detection to steer hot water production in an HP. 

In this paper, source and target domains are defined as follows:
\begin{align*}
\mathcal{D}_S &= \{\mathcal{X}_S, P_S(\mathcal{X})\}, \\
\mathcal{D}_T &= \{\mathcal{X}_T, P_T(\mathcal{X})\},
\end{align*}
where \( \mathcal{X}_S \) and \( \mathcal{X}_T \) denote the feature spaces \cite{Weiss2016} constructed from time-windowed input sequences, including $t_{\mathrm{mid}}$ and timestamp information. While the feature structure remains consistent across households, the marginal distributions \( P_S(\mathcal{X}) \) and \( P_T(\mathcal{X}) \) differ due to variations in the households' demand behavior.

Corresponding tasks are defined as:
\begin{align*}
\mathcal{T}_S &= \{\mathcal{Y}_S, f_S(\cdot)\}, \\
\mathcal{T}_T &= \{\mathcal{Y}_T, f_T(\cdot)\},
\end{align*}
where \( \mathcal{Y} \) is the label space representing future $t_{\mathrm{mid}}$ values, and \( f_S(\cdot) \) and \( f_T(\cdot) \) are the predictive functions for the source and target domains, respectively. 

TL aims to improve \( f_T(\cdot) \) by leveraging knowledge from \( \mathcal{D}_S \) and \( \mathcal{T}_S \), even when \( \mathcal{D}_S \neq \mathcal{D}_T \) \cite{Weiss2016}, which is often the case due to behavioral differences between households.

By design, DELTAiF is initiated by randomly selecting one household dataset as the source domain for pre-training the LSTM model with the initial hyperparameters described in Table \ref{tab:hyperparams}. Before training, all relevant household datasets are prepared for ML through rigorous handling of missing values and outliers.

\begin{table}
\centering
\caption{Hyperparameters used in LSTM model training}\label{tab:hyperparams}
\begin{tabular}{lll}
\toprule
\textbf{Model} & \textbf{Hyperparameters} & \textbf{Values} \\
\midrule
\multirow{5}{*}{\textbf{LSTM}}
 & LSTM units & 50 \\
 & Optimizer & Adam \\
 & Loss function & MAE \\
 & Epochs & 50 \\
 & Batch size & 72 \\
\midrule
\textbf{Response (Target)} & Predicted variable & $t_{\mathrm{mid}}$ \\
\bottomrule
\end{tabular}
\end{table}

The pretrained model is then fine-tuned on each of the remaining (n-1) household datasets, allowing the model to adapt to household-specific consumption patterns while retaining knowledge from the initial household. This process is repeated iteratively, such that each household is used once as the source domain. The goal is to evaluate transfer performance across all source-target combinations and identify the household that serves as the most effective source for knowledge transfer. The model performance is evaluated using widely recognized metrics in the demand forecasting literature, including MAPE, RMSE, and \( R^2 \) scores \cite{Nti2020}. In the last stage, evaluation results are documented and used to compare all pre-training scenarios, allowing the selection of the optimal source domain.

The second component of the DELTAiF pipeline fine-tunes the optimal pretrained model to forecast $t_\mathrm{mid}$ values for each of the target households. Based on these predictions, iForest is applied to identify anomalous drops in $t_\mathrm{mid}$, which are indicative of expected shower events. The detected events are then used to construct a household-specific consumption calendar that captures the frequency and temporal distribution of the expected weekly demand for hot water.

The integration of TL, LSTM-based forecasting, and iForest within the DELTAiF framework enables both accurate temperature prediction and adaptive hot water production at the household level. By learning household-specific consumption patterns, the system allows hot water production to be scheduled in a more demand-driven manner, thereby improving energy efficiency while maintaining user comfort.

\section{Implementation and evaluation} \label{sec:implementation}
This section describes the real-world HP installation datasets used to validate DELTAiF, including the different pre-processing phases performed in preparation for the implementation step.

\subsection{Experimental environment}
Table \ref{tab:environment} presents the computational environment used to implement DELTAiF. This table shows the specifications for the local machine used for initial pre-processing tasks and the server instance used for ML modeling, including the Graphics processing unit (GPU) configuration.

\begin{table}
\centering
\caption{Computational and environment specifications}\label{tab:environment}
\resizebox{8.4 cm}{!}{
\begin{tabular}{lll}
\toprule
\textbf{} & \textbf{Local environment} & \textbf{Server} \\
\midrule
\textbf{CPU} &Apple M1 Pro &Intel Core i9-9900X CPU \\
\textbf{GPU} & - &Nvidia GeForce RTX 2080, 48GB GDDR6\\
\textbf{RAM} & 16GB DDR4 &64GB DDR4\\
\bottomrule
\end{tabular}
}
\end{table}

\subsection{HPs real-installation datasets}
DELTAiF framework was evaluated on six real HP installations in residential units across Sweden. The collected data represent temperature recordings from the mid-tank sensor, where measurements are recorded every second or when any change occurs. The data cover recordings from 2022/2023 with a granularity of one second. The household datasets were carefully selected to include at least one year of data covering all seasons. The selected households exhibit various consumption patterns while operating on the same software platform. Table \ref{tab:dataset_info} describes start and end dates for train and test datasets per household and shows the size of the data after the split. The split approach ensured the train data covers full annual cycles, allowing for reliable model evaluation while reserving sufficient data for testing.

\begin{table*}
\centering
\caption{Information of households dataset}
\begin{tabular}{lcccccc}
\toprule
\textbf{Household} & 
\multicolumn{3}{c}{\textbf{Training and validation dataset}} & 
\multicolumn{3}{c}{\textbf{Test dataset}} \\
\cmidrule(lr){2-4} \cmidrule(lr){5-7}
& Start date & End date & Size & Start date & End date & Size \\
\midrule
\textbf{Household 1} & 14.10.2022 & 31.10.2023 & 521112 & 31.10.2023 & 24.11.2023 & 35384 \\
\textbf{Household 2} & 27.10.2022 & 31.10.2023& 530846 & 31.10.2023 & 24.11.2023 & 35384 \\
\textbf{Household 3} & 24.10.2022 & 31.10.2023 & 429676 & 31.10.2023 & 24.11.2023 & 35377 \\
\textbf{Household 4} & 03.11.2022 & 31.10.2023 & 429744 & 31.10.2023 & 24.11.2023 & 35385 \\
\textbf{Household 5} & 09.12.2022 & 31.10.2023 & 468228 & 31.10.2023 & 24.11.2023 & 35381 \\
\textbf{Household 6} & 29.10.2022 & 31.10.2023 & 429763 & 31.10.2023 & 24.11.2023 & 35385 \\
\bottomrule
\end{tabular}
\label{tab:dataset_info}
\end{table*}

In preparation for the application of DELTAiF, we conducted a thorough quality assessment using statistical and visualization tools. The results uncovered a key challenge: missing timestamps. This is caused by the method used to collect and store the sensor measurements data, where values are recorded only when changes occur, enabling the presence of missing timestamps in the time series. To address this problem, we restored missing timestamps using the forward-fill approach, which propagates the last valid observation forward \cite{Ribeiro2022}. Furthermore, we carried out an outlier detection and handling exercise to identify technical faults and false sensor readings. Outliers were discarded before the ML modeling phase. In addition, the data was then resampled to a consistent 1-minute granularity to ensure continuity of the time series data.

\section{Results} \label{sec:results}

\subsection{Cross-household transfer learning}

To evaluate the effectiveness of the proposed DELTAiF framework, it was implemented using real-world sensor data collected from six residential HP installations. The implementation was structured to validate the scalability of the framework across varying household consumption behaviors. The pipeline was executed such that each household served as both a source and a target domain. 

The results demonstrate asymmetric success in the performance of households datasets, particularly in the results for Households 1 and 5, as shown in Tables \ref{tab:training_hh1} and \ref{tab:training_hh5}. In Table \ref{tab:training_hh1}, Household 1 is used as a source domain (training) and the other household datasets are used as target domains. Household 1 pretrained model shows high generalization capabilities when used for the initial training phase, suggesting that it can capture hot water usage patterns across various household behaviors. However, the same household proves irregular consumption patterns when predicted, with RMSE greater than 1.2 in all six scenarios. On the contrary, Household 5 as a source domain performs consistently well as shown in Table \ref{tab:training_hh5}, where it achieves on average high \( R^2 \) values across all target households (0.874-0.991) and low MAPE values (0.001-0.017). Additionally, reasonable RMSE values (0.462-1.303) demonstrate high predictability across all five training models. The training results on Household 4 data, shown in Table \ref{tab:training_hh4}, show moderately predictable behavior, with moderate \( R^2 \) values and low MAPE and RMSE values. Observing the results of Households 2, 3, and 6 in Tables \ref{tab:training_hh2}, \ref{tab:training_hh3}, and \ref{tab:training_hh6}, they are likely to have more consistent and regular hot water usage patterns.

\begin{table}
\caption{Evaluation Metrics by training on Household 1}
\label{tab:training_hh1}
\centering
\begin{tabular}{lcccc}
\toprule
\textbf{Household \#} & MAPE & RMSE & \( R^2 \) \\
\midrule
\textbf{Household 1 (training)} & 0.017 & 1.314 & 0.50 \\
\textbf{Household 2}  & 0.004 & 0.485 & 0.991 \\
\textbf{Household 3}  & 0.004 & 0.834 & 0.983 \\
\textbf{Household 4}  & 0.0007 & 0.758 & 0.870 \\
\textbf{Household 5}  & 0.002 & 0.250 & 0.996 \\
\textbf{Household 6}  & 0.004 & 0.852 & 0.966 \\
\bottomrule
\end{tabular}
\end{table}

\begin{table}
\caption{Evaluation Metrics by training on Household 2}
\label{tab:training_hh2}
\centering
\begin{tabular}{lcccc}
\toprule
\textbf{Household \#} & MAPE & RMSE & \( R^2 \) \\
\midrule
\textbf{Household 1} & 0.016 & 1.236 & 0.562 \\
\textbf{Household 2 (training)} & 0.003 & 0.41 & 0.99 \\
\textbf{Household 3} & 0.004 & 0.822 & 0.983 \\
\textbf{Household 4} & 0.0009 & 0.740 & 0.876 \\
\textbf{Household 5} & 0.002 & 0.265 & 0.996 \\
\textbf{Household 6} & 0.004 & 0.816 & 0.969 \\
\bottomrule
\end{tabular}
\end{table}

\begin{table}
\caption{Evaluation Metrics by training on Household 3}
\label{tab:training_hh3}
\centering
\begin{tabular}{lcccc}
\toprule
\textbf{Household \#} & MAPE & RMSE & \( R^2 \) \\
\midrule
\textbf{Household 1}  & 0.0184 & 1.398 & 0.440 \\
\textbf{Household 2}  & 0.006 & 0.526 & 0.989 \\
\textbf{Household 3 (training)}  & 0.004 & 0.760 & 0.986 \\
\textbf{Household 4} & 0.0007 & 0.767 & 0.867 \\
\textbf{Household 5} & 0.003 & 0.279 & 0.995 \\
\textbf{Household 6} & 0.003 & 0.790 & 0.971 \\
\bottomrule
\end{tabular}
\end{table}

\begin{table}
\caption{Evaluation Metrics by training on Household 4}
\label{tab:training_hh4}
\centering
\begin{tabular}{lcccc}
\toprule
\textbf{Household \#} & MAPE & RMSE & \( R^2 \) \\
\midrule
\textbf{Household 1} & 0.018 & 1.397 & 0.441 \\
\textbf{Household 2} & 0.004 & 0.509 & 0.990 \\
\textbf{Household 3} & 0.0038 & 0.759 & 0.986 \\
\textbf{Household 4 (training)} & 0.0007 & 0.756 & 0.87 \\
\textbf{Household 5} & 0.004 & 0.305 & 0.995 \\
\textbf{Household 6} & 0.004 & 0.832 & 0.968 \\
\bottomrule
\end{tabular}
\end{table}

\begin{table}
\caption{Evaluation Metrics by training on Household 5}
\label{tab:training_hh5}
\centering
\begin{tabular}{lcccc}
\toprule
\textbf{Household \#} & MAPE & RMSE & \( R^2 \) \\
\midrule
\textbf{Household 1} & 0.017 & 1.303 & 0.514 \\
\textbf{Household 2} & 0.003 & 0.462 & 0.991 \\
\textbf{Household 3} & 0.004 & 0.809 & 0.984 \\
\textbf{Household 4} & 0.001 & 0.746 & 0.874 \\
\textbf{Household 5 (training)} & 0.003 & 0.283 & 0.995 \\
\textbf{Household 6} & 0.005 & 0.811 & 0.969 \\
\bottomrule
\end{tabular}
\end{table}

\begin{table}
\caption{Evaluation Metrics by training on Household 6}
\label{tab:training_hh6}
\centering
\begin{tabular}{lcccc}
\toprule
\textbf{Household \#} & MAPE & RMSE & \( R^2 \) \\
\midrule
\textbf{Household 1} & 0.018 & 1.371 & 0.462 \\
\textbf{Household 2} & 0.003 & 0.453 & 0.992 \\
\textbf{Household 3} & 0.003 & 0.728 & 0.987 \\
\textbf{Household 4} & 0.001 & 0.752 & 0.872 \\
\textbf{Household 5} & 0.002 & 0.254 & 0.996 \\
\textbf{Household 6 (training)} & 0.0038 & 0.818 & 0.969 \\
\bottomrule
\end{tabular}
\end{table}

\begin{figure}
    \centering
    \includegraphics[width=1\linewidth]{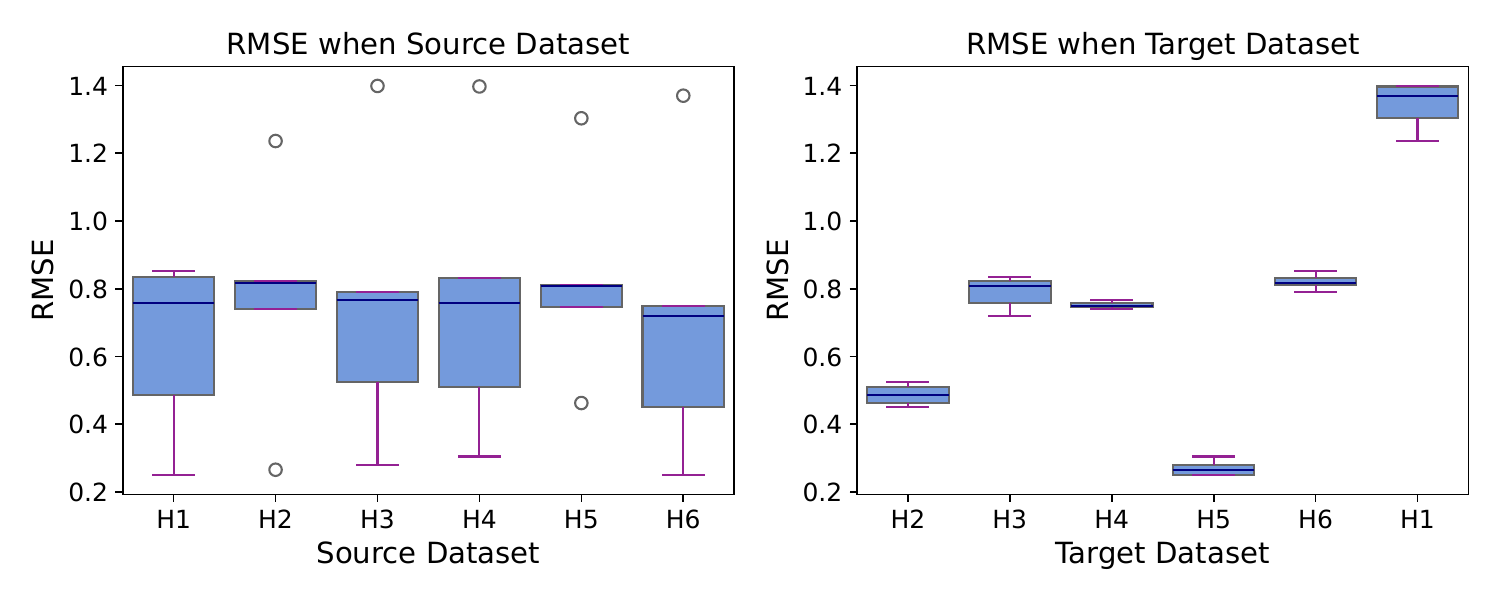}
    \caption{Overall RMSE for source and target datasets}
    \label{fig:rmse}
\end{figure}

Additionally, MAPE and RMSE distributions were examined for all six datasets to evaluate the suitability of household candidates as source and target domains. In Figure \ref{fig:rmse}, Households 5 show lower RMSE variability compared to other household datasets, which makes it a strong source and a predictable target. Therefore, such datasets are ideal candidates for developing generalized TL models.

\begin{figure}
    \centering
    \includegraphics[width=1\linewidth]{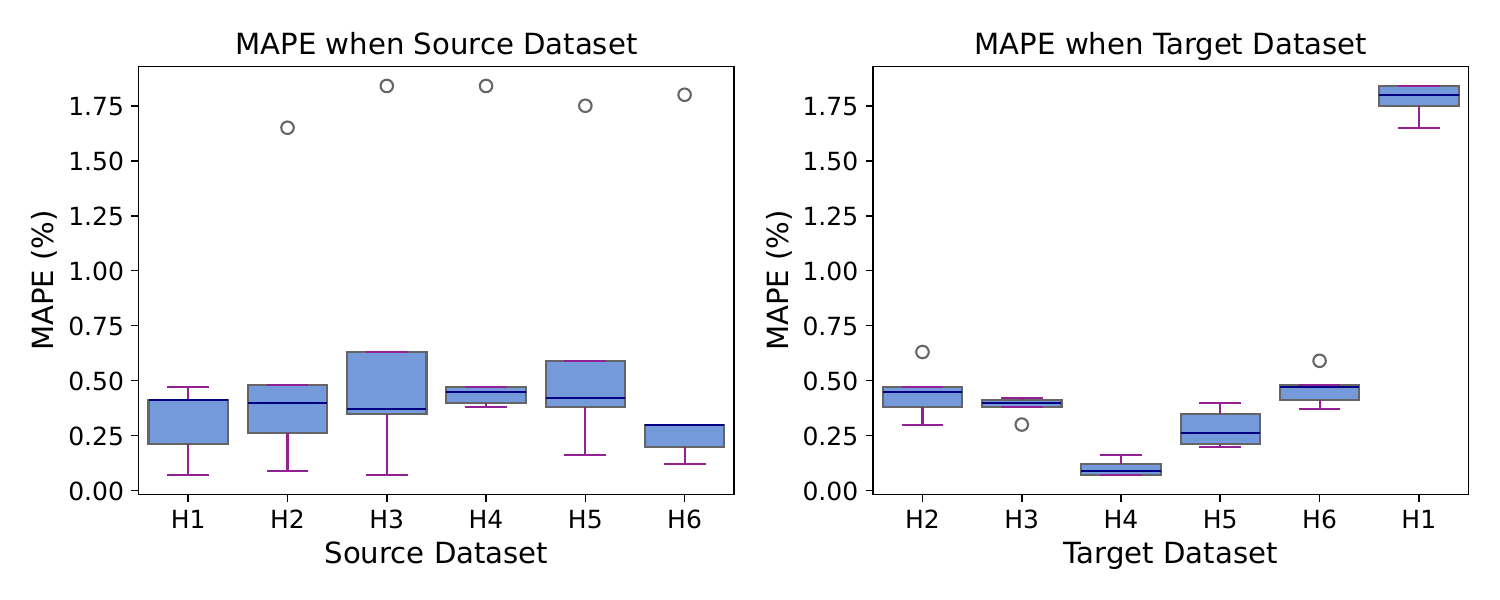}
    \caption{Overall MAPE for source and target datasets}
    \label{fig:mape}
\end{figure}

Similarly, in Figure \ref{fig:mape}, Household 4 as a source dataset demonstrates consistently low variability in MAPE values, indicating potential suitability for generalization. Other datasets, when used as a source, show relatively higher MAPE variability, suggesting that the model can struggle to generalize to different household consumption behaviors. However, Household 5 has the lowest mean MAPE among all source domains. This indicates that using Household 5 as a source domain yields the most accurate predictions on average.

Among all households, training on Household 5 delivered the lowest mean RMSE and MAPE across all target households, proving its suitability as the most effective source domain. This success can be attributed to Household 5 stable and regular consumption patterns, facilitating the model’s ability to generalize across different household behaviors. When used as a source domain, it demonstrates good potential to capture the general patterns of hot water temperature behavior, even for the most challenging target, Household 1. As a result, Household 5 comprises the most generalizable patterns of hot water consumption behavior in HPs, making it an ideal starting point for TL.

Regarding run time, training the model from scratch for each household data takes an average of 9 minutes per household, under the condition of the computational environment, resulting in a total runtime of 54 minutes for all six households. With TL, the model is initially trained on one household for 9 minutes and then fine-tuned on the data from the remaining five households in just 8.55 minutes. This reduced the total runtime and saved approximately 67\% of the total computation time. These results demonstrated the efficiency of TL in reducing training costs while effectively adapting the prediction model to each household consumption pattern.

\subsection{Construction of household weekly demand calendar}
\begin{figure*}
    \centering
    \begin{subfigure}[b]{0.4\textwidth}
        \includegraphics[width=\textwidth]{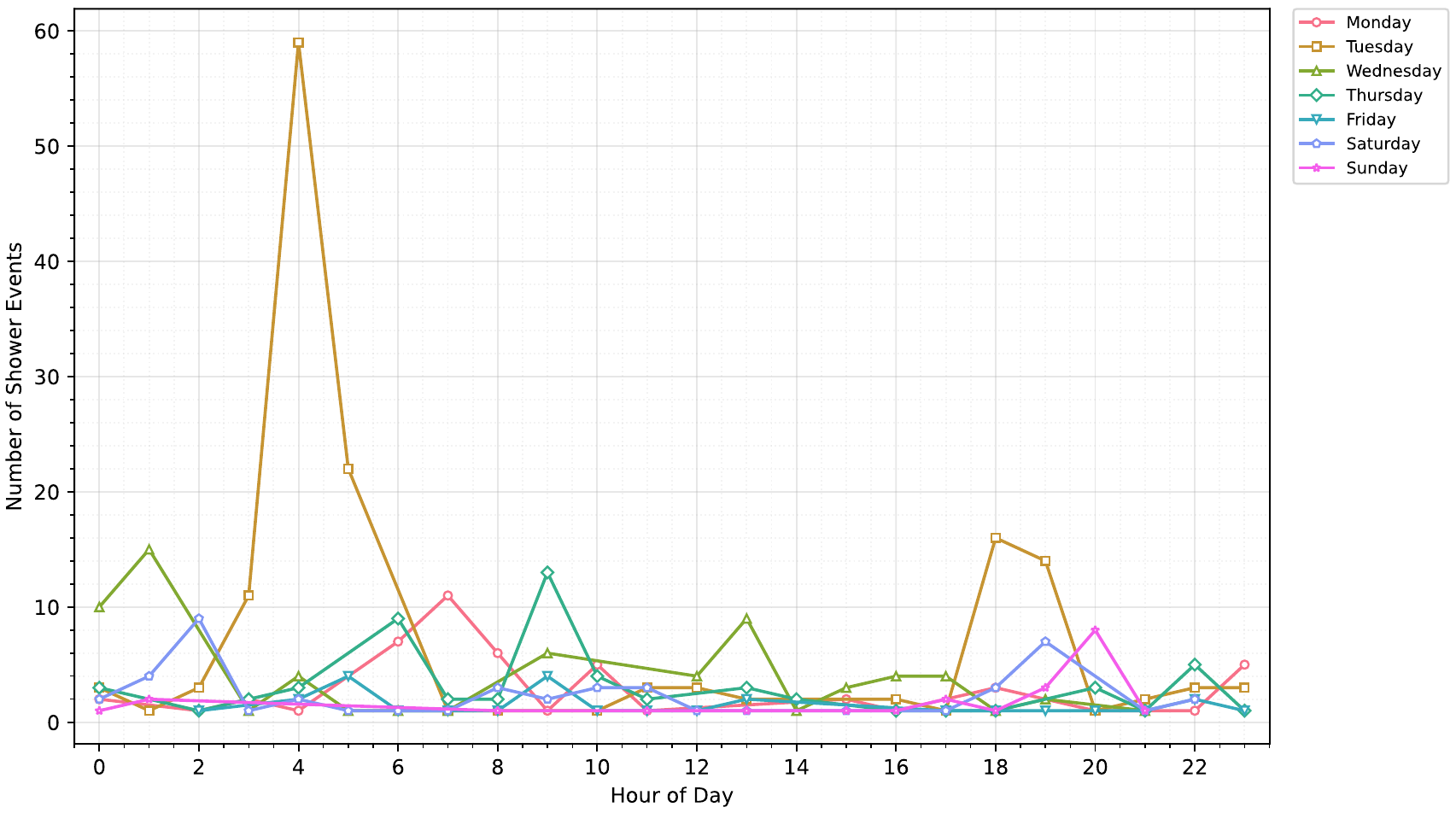}
        \caption{Household 1}
        \label{fig:household39}
    \end{subfigure}
    \hfill
    \begin{subfigure}[b]{0.4\textwidth}
        \includegraphics[width=\textwidth]{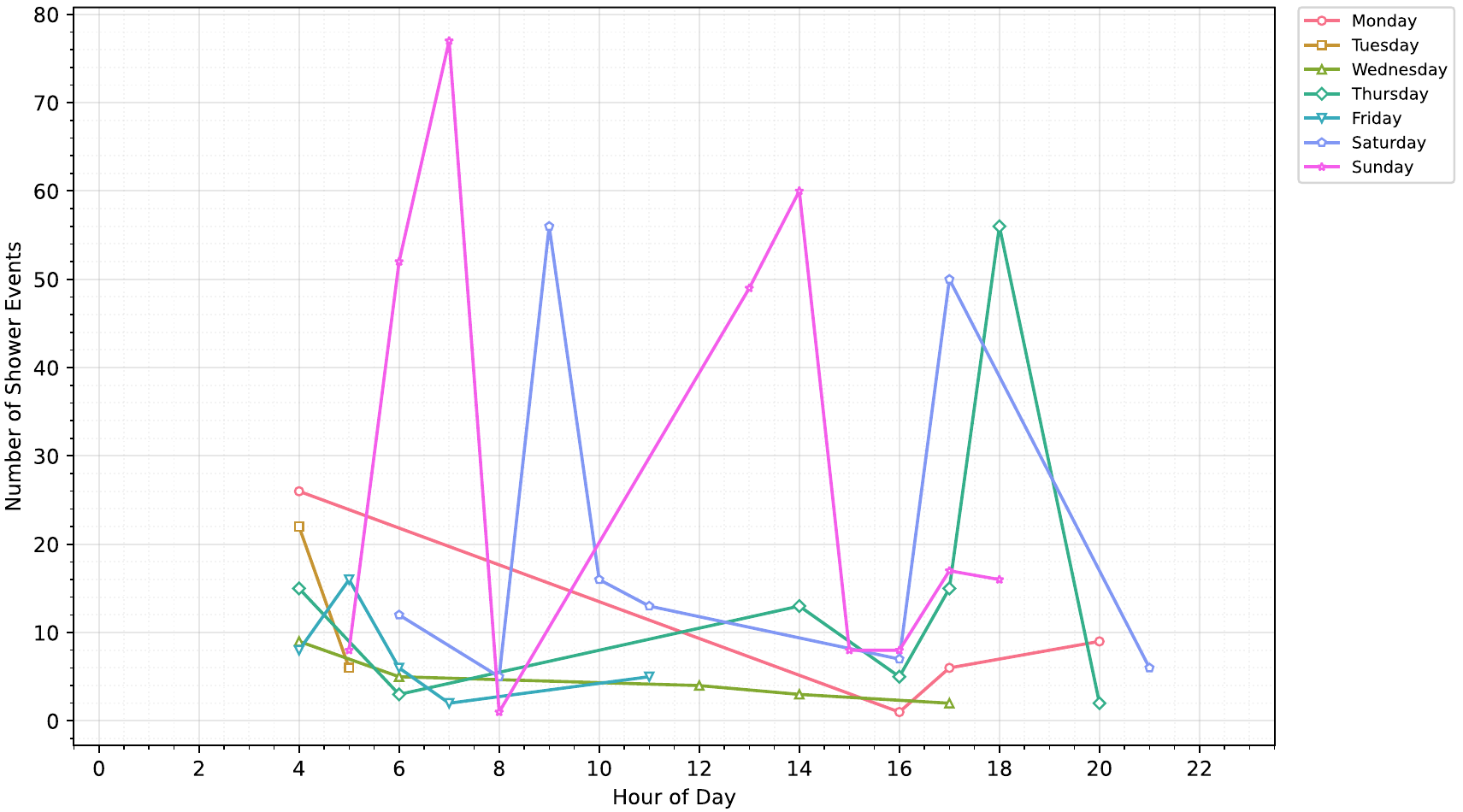}
        \caption{Household 2}
        \label{fig:household41}
    \end{subfigure}
    \begin{subfigure}[b]{0.4\textwidth}
        \includegraphics[width=\textwidth]{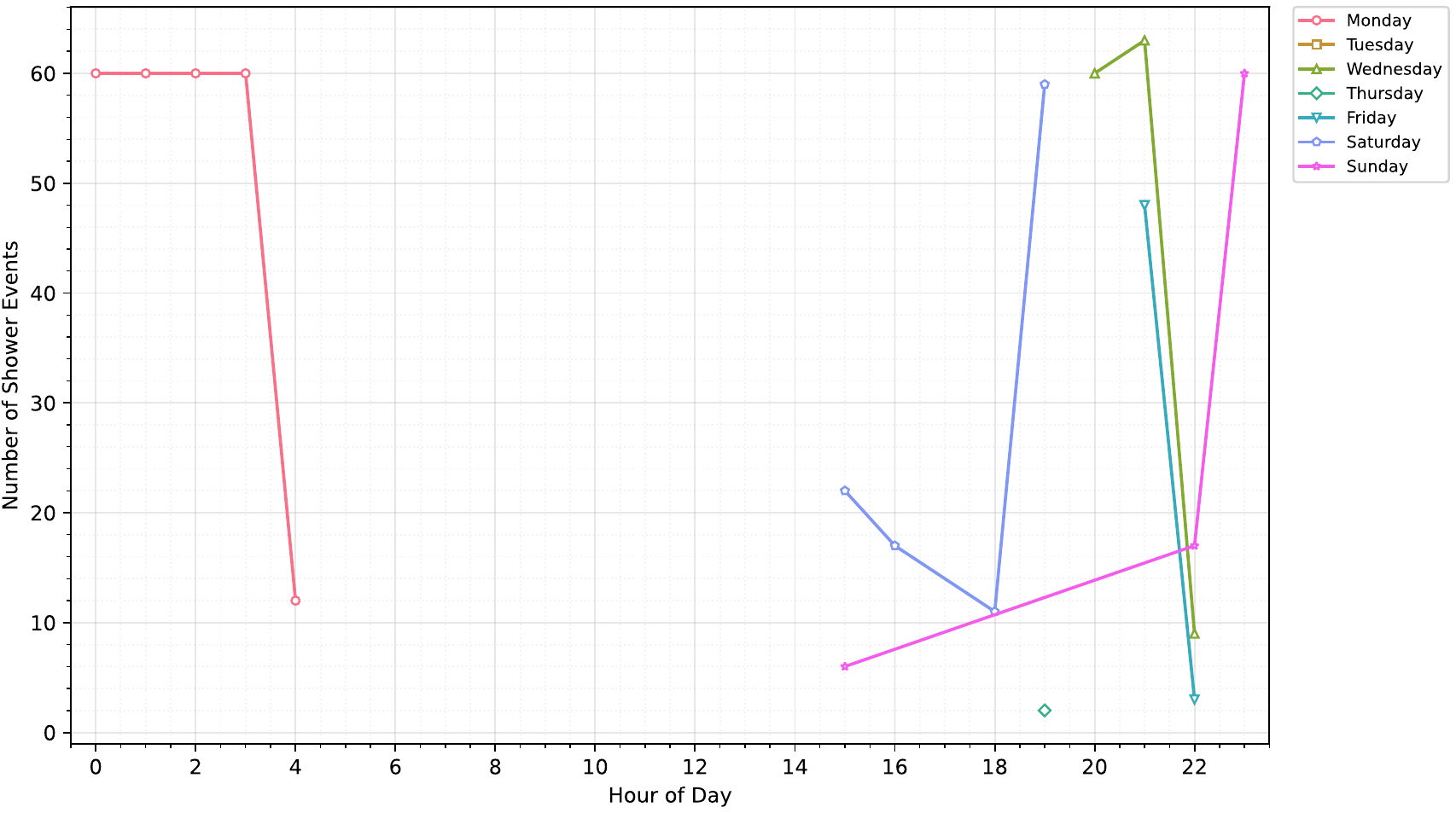}
        \caption{Household 3}
        \label{fig:household43}
    \end{subfigure}
    \hfill
    \begin{subfigure}[b]{0.4\textwidth}
        \includegraphics[width=\textwidth]{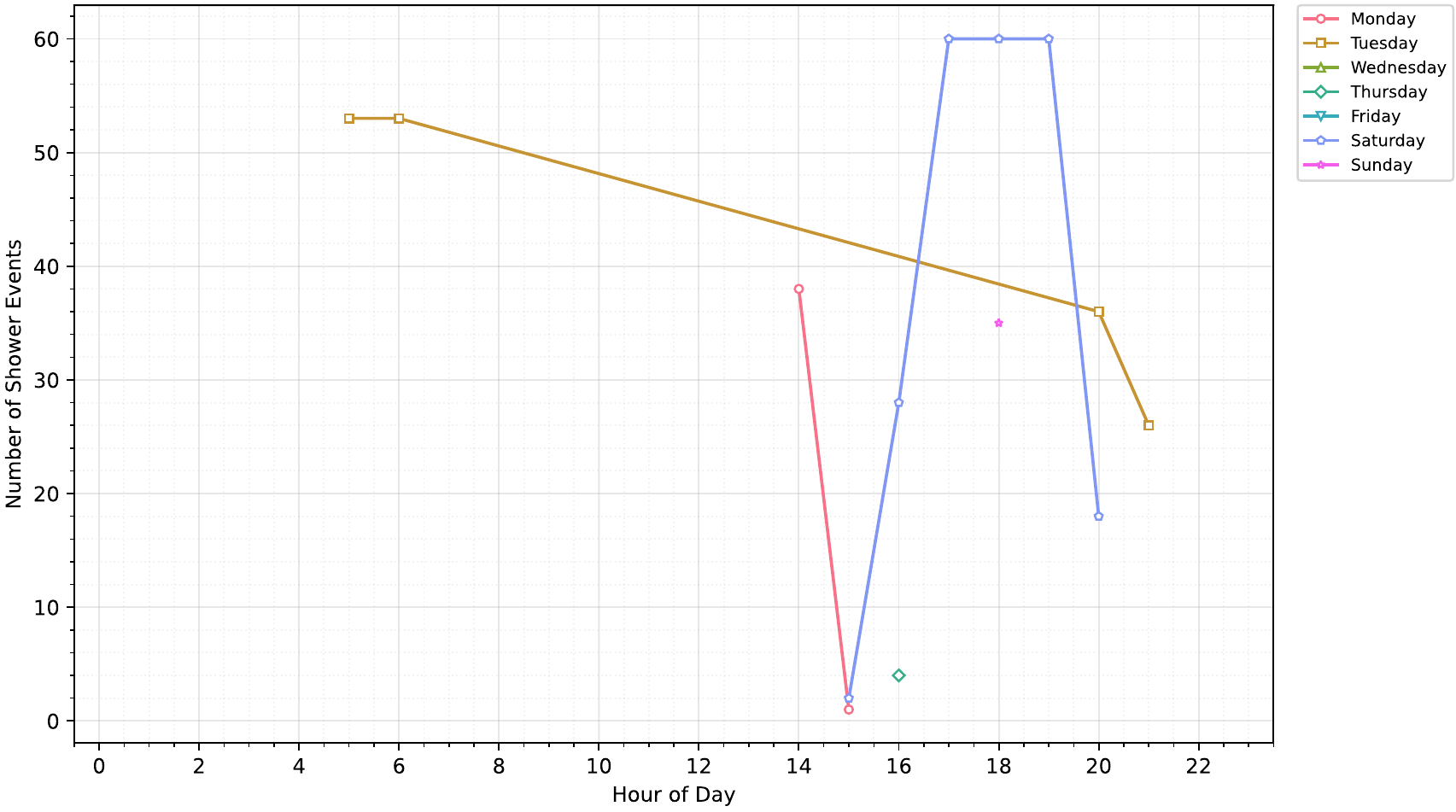}
        \caption{Household 4}
        \label{fig:household44}
    \end{subfigure}
    \begin{subfigure}[b]{0.4\textwidth}
        \centering
        \includegraphics[width=\textwidth]{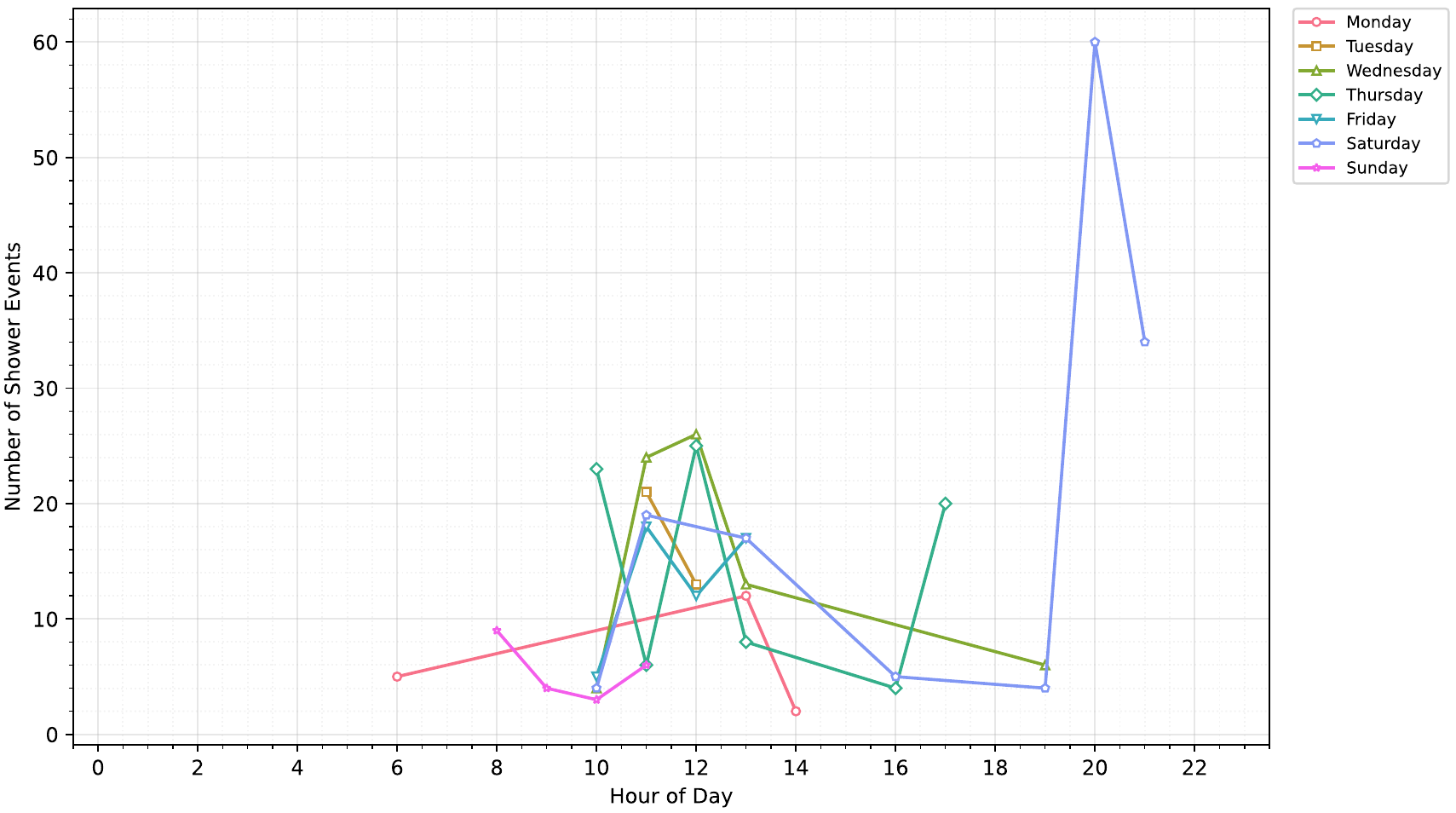}
        \caption{Household 6}
        \label{fig:household51}
    \end{subfigure}

    \caption{Predicted weekly hot water consumption calendar for target households, contamination rate = 0.02}
    \label{fig:consumptioncalendar}
\end{figure*}

After training the LSTM model on Household 5 and fine-tuning it on the remaining households, DELTAiF proceeded with its second component, constructing the weekly demand calendar, based on the fine-tuned predictions. Thereafter, the fine-tuned model's predictions were passed to the iForest algorithm to detect shower events. The pipeline can remember and processes time stamp information to generate shower event calendars and analyze weekly trends, which is enabled by design. The anomaly threshold known as the contamination rate, was selected to be 2\%. This threshold was selected based on empirical analysis of household water consumption patterns, where infrequent events such as showers typically fall within the top 2\% of detected anomalies. Throughout the process, predicted weekly activity was analyzed and visualized as shown in Figure \ref{fig:consumptioncalendar}.

Analysis of Household 1 results illustrated in Figure \ref{fig:consumptioncalendar} shows peaks of shower events on Tuesdays at 04:00 and 18:00 hours. Weekday shower patterns demonstrate greater variability compared to weekends, with Wednesday showing consistent low-level activity throughout the day. Monday and Thursday exhibit moderate usage patterns with peaks at 07:00-08:00 and 09:00-10:00 hours, respectively. 

As for Household 2, the weekly trend displays multiple high-intensity periods throughout the day, including Sunday mornings. The pattern shows three distinct major peaks: Sunday morning, Sunday afternoon, and Saturday evening. This suggests that Household 2 hot water usage is concentrated on the weekends. Weekday usage is more moderate but still substantial, with consistent activity levels throughout weekdays.

Household 3 demonstrates an interesting shift toward late-day usage, with peak activity occurring between 20:00 and 22:00 hours, particularly on Wednesdays. Monday mornings show consistently high usage followed by a sharp decline, suggesting this might be a household with night shift workers or late-night activities.

The pattern in Household 4 is characterized by sustained stable periods rather than sharp peaks. Saturday evenings, between 16:00-20:00 hours, maintain a consistent level of high consumption of hot water, while Tuesday mornings show moderate usage. The extended periods of high usage rather than sharp peaks may suggest this household might have occupants with scheduled group activities.

Household 6 has more expected pattern with a clear evening preference, particularly on Saturdays at 20:00 hours. The majority of shower activity is concentrated in the late morning to evening hours (10:00-22:00), with minimal early morning usage. This pattern may suggest residential regular working hours, where occupants follow more traditional daily schedules.

The different consumption behaviors among the selected households, such as Household 1 Tuesday morning peak, Household 2 multiple daily peaks, Household 3 late-day preferences, and Household 5 conventional evening peaks, stretch the importance of consumption for adaptive hot water production. These varied patterns justify the need for DELTAiF framework, where pre-training on selected household data followed by fine-tuning on other households can capture both common and unique usage patterns. The identified patterns are key information that can be transformed and shared with the HP operating system for steering the production of hot water in a demand-driven manner in households.

\section{Discussion}\label{sec:discussion}
The evaluation results show DELTAiF efficiency in reducing computational burden, which makes it particularly well-suited for large-scale cloud deployments. This solution is especially relevant for cloud-connected systems managing thousands of HP installations, where scalability is a real challenge. The 67\% reduction in training time, compared to a total of 54 minutes for traditional training, represents a significant contribution to enhancing scalability. The promising advantages of TL in leveraging knowledge from existing installations to train individual models for each household provide a sustainable and automated approach to scaling. Therefore, our proposed framework is particularly valuable in the case of growing connected HPs, offering a solution that balances cost efficiency with prediction accuracy.

Furthermore, the results reveal interesting insights into the characteristics that make certain households more suitable as source domains for TL, particularly in the case of Household 5, where its success can be attributed to its typical and consistent consumption patterns. This kind of consumption behavior captures fundamental features of hot water usage that generalize well across different household behaviors. On the other hand, households with irregular consumption patterns, such as Household 1, prove that unpredictable usage patterns can significantly impact the ML model performance.

\section{Threats to validity}\label{sec:threats}

To assess the proposed framework, we evaluated it for forecasting hot water consumption use cases with LSTM models. However, several constraints should be considered. First and foremost, the framework's effectiveness depends on the quality and consistency of source domain data, where irregular consumption patterns can impact prediction accuracy. Therefore, the selection of a source domain for pre-training is a key decision. Furthermore, the successful implementation of DELTAiF assumes relatively stable household consumption patterns over time, so the presence of long-term behavioral changes in the data might require a model update to achieve the desired performance. While no explicit labeled ground truth was available, iForest with a fixed contamination rate of 2\% showed to detect shower events effectively. However, the fixed contamination rate of 2\%  may not be optimal for all household consumption patterns. Additionally, to validate the detected shower events, a manual inspection of consumption patterns across multiple households was conducted. The observations confirm that the majority of anomalies identified by iForest corresponded to sharp hot water withdrawals. Finally, the current evaluation of DELTAiF is limited to residential HP installations, and in the case of adaptation to industrial settings, modifications to the framework's design may be needed.

\section {Conclusion}\label{sec:conclusion}
TL has been deployed in various demand forecasting applications, where it has shown promising results. This paper proposes DELTAiF, a framework that leverages TL to identify the optimal source domain and predict hot water consumption in domestic households. The proposed framework offers a scalable solution for modeling hot water consumption across multiple HP installations while significantly reducing computational overhead. DELTAiF evaluation is carried out using six household installation data, where the prediction performance is measured using MAPE and RMSE error metrics. The evaluation results demonstrate the efficiency of our solution, particularly in that households have distinct consumption patterns from regular weekday/weekend cycles to unique timing preferences. The DELTAiF framework successfully captures these variations while maintaining high prediction accuracy. We observe that efficient performance is achieved when using households with stable consumption patterns as the source domain for TL.
 
This framework applies to scenarios where regular and irregular real-world domestic routines exist. DELTAiF presents an opportunity to enhance and complement traditional control methods based on fixed thresholds and setpoint values. By enabling adaptive, data-driven decision-making, ML can allow HP systems to dynamically adjust to consumption patterns, improving operational efficiency and reducing energy waste.

Beyond the immediate application of DELTAiF to HP systems, this work has broader implications for energy systems. The significant reduction in computational requirements makes this approach particularly valuable for large-scale deployments where individual model training is expensive. Future research could expand this methodology to other aspects of building energy management and investigate automated source domain selection based on consumption pattern stability.

\section*{Acknowledgments} 
This work was supported by the Swedish Knowledge Foundation via the KKS Industrial Research School program “EXACT - Excellence in Advancing for a Circular Transition” (grant number 20220134).
\balance
\bibliographystyle{IEEEtran}

\bibliography{refs}

\end{document}